\documentclass{article}
\usepackage{spconf,amsmath,amssymb,amsfonts,graphicx}
\usepackage{microtype}
\usepackage{cite,textcomp,xcolor,booktabs,array,multirow,threeparttable}
\usepackage[normalem]{ulem}
\usepackage{xurl}
\usepackage[hidelinks]{hyperref}
\usepackage{orcidlink}

\newcommand{\best}[1]{\textbf{#1}}
\newcommand{\snd}[1]{\uline{#1}}
\newcommand{\gray}[1]{\textcolor{gray}{#1}}

\title{Parameter-Efficient Adaptation of Self-Supervised Vision Backbones
Survives Distribution Shift in Radar Micro-Doppler HAR}
\name{\shortstack{
Keryan Chelouche$^{\star \dagger}$\,\orcidlink{0009-0002-1692-1827},
Petr Dobias$^{\star}$\,\orcidlink{0000-0003-2969-5259},\\
Nistor Grozavu$^{\star}$\,\orcidlink{0000-0001-7502-8022},
Olivier Romain$^{\star}$\,\orcidlink{0000-0002-2172-1865}%
\thanks{This work was supported by the French National Research
Agency (ANR) under grant 24-LCV2-0020-01 /SmartGaitLab.}
}}

\address{
$^{\star}$ETIS UMR 8051, CY Cergy Paris Universite, ENSEA, CNRS,
F-95000 Cergy, France\\
$^{\dagger}$Fondation Léonie Chaptal, F-95200 Sarcelles, France
}
\begin{document}
\ninept
\maketitle
\begin{abstract}
Micro-Doppler radar enables contactless, privacy-preserving 
human activity recognition (HAR), but deployment robustness 
remains underexplored. We compared full fine-tuning, frozen probing, 
and parameter-efficient fine-tuning (PEFT) using subject-disjoint
in-distribution cross-validation and few-shot cross-age, cross-orientation, 
and cross-dataset transfer. Full fine-tuning suffered negative cross-dataset 
transfer, while frozen features under-adapted on cross-orientation. 
Only PEFT avoided both failures. We paired DINOv3 with all-layer 
principal-singular-direction adaptation (PiSSA), to our knowledge 
the first application of a self-supervised image foundation backbone 
to radar micro-Doppler HAR. DINOv3-PiSSA ranked first among PEFT methods in every evaluated setting. 
Under our protocol, its Small variant matched the in-distribution accuracy of the prior Base-size PEFT state of the art using $4.5\times$ fewer total parameters, with only 3\% trainable. 
It surpassed a size-matched variant of that baseline in every evaluated setting,
with transfer gains increasing with shift severity: 2.1, 6.6, and 8.9 percentage points, respectively.
Ablations attributed improvements to both pretraining and adapter recipe, 
with the pretraining contributing most under compound shift.
Code and protocol are available at
\url{https://github.com/KeryanChelouche/udoppler-ssl}.
\end{abstract}
\section{Introduction}
Falls are the second leading cause of unintentional-injury death
worldwide~\cite{who_falls}. Radar offers contactless, privacy-preserving sensing
for ambient activity monitoring and fall detection, with robustness to lighting
and occlusion~\cite{gurbuz2019,radarreview}. Human motion produces a
time--frequency signature, the micro-Doppler spectrogram, which can be treated
as an image. This representation has enabled the adoption of computer-vision
classifiers, from CNN and CNN--LSTM architectures~\cite{kim2016,li2020,diatradharnet}
to pretrained vision transformers. SelaFD~\cite{selafd}, the current state of the
art on the public University of Glasgow benchmark~\cite{glasgow}, reported 96.6\% accuracy.
It adapted a supervised ImageNet ViT-B using attention LoRA and serial--parallel
bottleneck adapters.

However, these accuracies provide little evidence of robustness to deployment
shifts. Evaluation in this literature is predominantly in-distribution, with
training and test data sharing radar, environment, and sensing geometry.
Common protocols also use a single random split and training seed, without
guaranteeing subject disjointness~\cite{kim2016,li2020,diatradharnet,selafd}.
The effects of seed variability~\cite{picard2021,bouthillier2021}, distribution
shift~\cite{recht2019,torralba2011}, and record-wise leakage~\cite{saeb2017}
have been documented beyond radar recognition.

Deployment introduces several sources of shift. A fall detector trained on
young laboratory volunteers must serve older users whose gait and fall dynamics
differ. The Glasgow dataset's own authors reported accuracy losses of roughly
10--20 percentage points when classifiers trained on younger subjects were
tested on older adults in care settings~\cite{shah2022}. Sensing geometry also
changes when users move along arbitrary trajectories, because micro-Doppler
encodes only radial velocity~\cite{chen2003}. Changes in hardware, signal
processing, environment, and population introduce a compound shift.

The candidate adaptation strategies span full fine-tuning, frozen linear
probing, and parameter-efficient fine-tuning (PEFT). Full fine-tuning is the
field's incumbent, while self-supervised foundation models have made frozen
probing competitive on natural images~\cite{caron2021,oquab2024}. PEFT occupies
the middle ground by adapting a small fraction of parameters. Concurrent work
on WiFi micro-Doppler found that frozen DINOv2 features narrowed an external
cross-dataset gap where supervised features and input-level adaptation fell
short~\cite{wifidino}. That study evaluated frozen encoders without comparing
fine-tuned or PEFT variants. The relative effectiveness of these adaptation
strategies under realistic shifts in radar micro-Doppler HAR therefore
remained, to our knowledge, unexamined. With limited labeled radar data, adaptation must balance learning 
task-specific features with retaining transferable pretrained representations. 
We hypothesized that combining self-supervised image pretraining with 
parameter-efficient adaptation would improve this balance 
under deployment shifts. We therefore asked two research questions.

\textbf{RQ1:} Which adaptation regime best
preserves in-distribution performance and few-shot transfer under
cross-age, cross-orien\-tation, and cross-dataset shifts? \textbf{RQ2:} Can
self-supervised image backbones improve on supervised ImageNet PEFT, and how
do pretraining and adapter recipe contribute to transfer performance?

\begin{figure}[t]
\centerline{\includegraphics[width=\columnwidth]{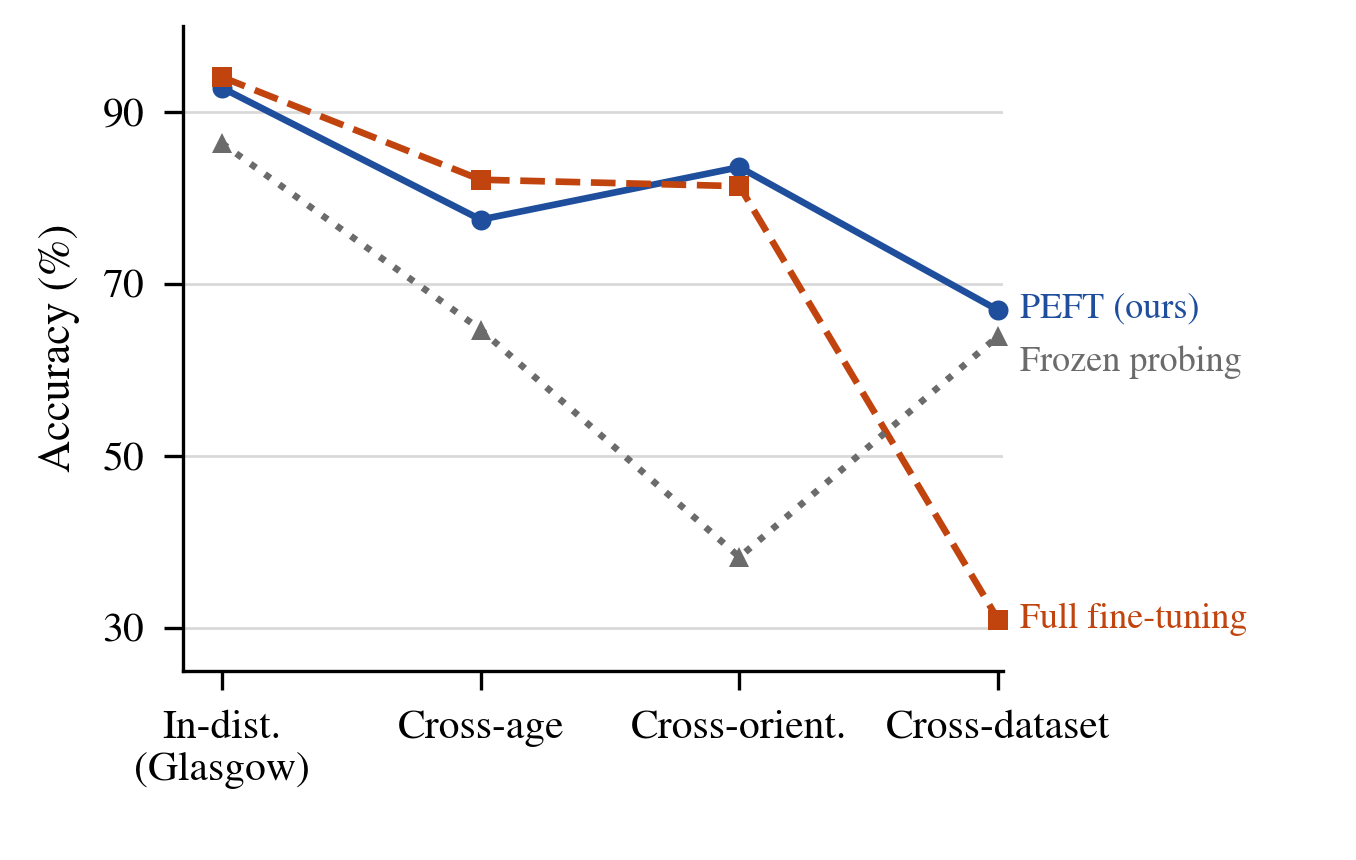}}
\caption{Three adaptation strategies using DINOv3-S. In-distribution accuracy
was evaluated on Glasgow. Transfer under the three shifts used ten labeled
target examples per class, with subject-disjoint support and test sets.
Full fine-tuning collapsed on cross-dataset, while frozen probing under-adapted
on cross-orien\-tation. Only PEFT avoided both failure regimes.}
\label{fig:teaser}
\end{figure}

Our subject-disjoint evaluation revealed contrasting failure regimes for full
fine-tuning and frozen probing. Only PEFT avoided both across the three shifts
(Fig.~\ref{fig:teaser}). Building on this finding, we paired DINOv3 with
all-layer PiSSA. To our knowledge, this was the first application of a
self-supervised image foundation backbone to radar micro-Doppler HAR.
DINOv3-PiSSA ranked first among PEFT methods in every evaluated setting,
training only 3\% of backbone parameters at Small size. Ablations showed that
both components improved transfer, with the backbone contributing most under
compound shift.
\section{Method}
\subsection{Self-supervised backbone}\label{sec:backbone}
We used DINOv3~\cite{dinov3} ViT-S/16 and ViT-B/16, with 384- and
768-dimensional embeddings, respectively. Both were pretrained on natural
images without labels. Prior micro-Doppler methods instead adapted supervised
ImageNet encoders. We hypothesized that label-free features were less committed
to ImageNet's semantic categories and would transfer better to micro-Doppler
spectrograms. All ViT-based classifiers used a single linear readout of the CLS embedding,
including fine-tuned heads, frozen probes, and few-shot target readout.
\subsection{Principal-direction parameter-efficient adaptation}\label{sec:pissa}
For a pretrained weight $W_{\mathrm{pre}}\in\mathbb{R}^{d\times k}$, we used
\begin{equation}
W = W_{\mathrm{res}} + sBA.
\label{eq:lora}
\end{equation}
Here, $d$ and $k$ are the output and input dimensions,
$B\in\mathbb{R}^{d\times r}$, and $A\in\mathbb{R}^{r\times k}$. The adapter
rank satisfies $r\ll\min(d,k)$, and $s=\alpha/r$ sets the scale through the
coefficient $\alpha$~\cite{lora}. Only $A$, $B$, and the classification head were
trained. Standard LoRA freezes $W_{\mathrm{res}}=W_{\mathrm{pre}}$ and initializes
$B=0$. Our recipe changed both layer coverage and initialization.

\emph{Coverage.} We placed adapters on every attention and multilayer perceptron
(MLP) projection ($q,k,v,o$, up, down), rather than attention alone. We used
$r=8$ and $\alpha=16$, yielding 0.67/1.33 million trainable parameters for
Small/Base, respectively (3.0\%/1.5\% of total model parameters).

\emph{Initialization.} Following PiSSA~\cite{pissa}, we decomposed each
pretrained weight as $W_{\mathrm{pre}}=U\Sigma V^\top$.
Let $U_r\in\mathbb{R}^{d\times r}$ and
$V_r\in\mathbb{R}^{k\times r}$ contain the left and right singular
vectors associated with the $r$ largest singular values, collected
in the diagonal matrix $\Sigma_r\in\mathbb{R}^{r\times r}$.
We initialized
\begin{equation}
B_0=s^{-1/2}U_r\Sigma_r^{1/2},
\qquad
A_0=s^{-1/2}\Sigma_r^{1/2}V_r^\top,
\label{eq:pissa_init}
\end{equation}
and froze the residual
\begin{equation}
\begin{aligned}
W_{\mathrm{res}}
&= W_{\mathrm{pre}}-sB_0A_0\\
&= W_{\mathrm{pre}}-U_r\Sigma_rV_r^\top.
\end{aligned}
\label{eq:pissa_residual}
\end{equation}
Thus, $W_{\mathrm{res}}+sB_0A_0=W_{\mathrm{pre}}$: initialization
preserved the pretrained linear transformation. 
\begin{table*}[t]
\caption{In-Distribution Accuracy on Glasgow and MAD (\%)}
\label{tab:indist}
\centering
\begin{threeparttable}
\begin{tabular}{lllcc}
\toprule
Method & Trainable & Total & Glasgow & MAD \\
\midrule
\multicolumn{5}{l}{\textit{Full fine-tuning}}\\
\gray{ResNet50} & \gray{23.52\,M (100\%)} & \gray{23.52\,M} & \gray{94.57 (1.08)} & \gray{97.94 (0.59)}\\
\gray{DINOv3-S (full FT)} & \gray{21.60\,M (100\%)} & \gray{21.60\,M} & \gray{94.04 (0.78)} & \gray{97.86 (0.61)}\\
\addlinespace
\multicolumn{5}{l}{\textit{Linear probing on frozen models}}\\
ViT-S (IN)   & $\sim$0 (probe only) & 21.67\,M & 87.16 (1.58) & 63.51 (1.24)\\
ViT-B (IN)   & $\sim$0 (probe only) & 85.80\,M & 88.26 (2.03) & 67.36 (1.88)\\
DINOv3-S     & $\sim$0 (probe only) & 21.60\,M & 86.35 (1.52) & 73.34 (1.00)\\
DINOv3-B     & $\sim$0 (probe only) & 85.66\,M & 87.12 (1.28) & 74.83 (0.93)\\
\addlinespace
\multicolumn{5}{l}{\textit{PEFT-based}}\\
ViT-S + LoRA & 0.22\,M (1.02\%)  & 21.89\,M  & 91.92 (1.20) & 94.14 (0.73)\\
SelaFD-S$^{*}$ & 3.63\,M (14.35\%)  & 25.29\,M  & 90.77 (1.19) & 95.96 (0.66)\\
SelaFD$^{*}$   & 14.34\,M (14.32\%) & 100.13\,M & 92.06 (0.72) & 96.82 (0.73)\\
DINOv3-PiSSA-S & 0.67\,M (2.99\%)  & 22.26\,M  & \snd{92.79 (1.37)} & \snd{97.43 (0.47)}\\
DINOv3-PiSSA-B & 1.33\,M (1.53\%)  & 86.99\,M  & \best{93.85 (1.15)} & \best{97.63 (0.51)}\\
\bottomrule
\end{tabular}
\begin{tablenotes}[flushleft]\normalsize
\item Mean accuracy (standard deviation) across five subject-disjoint folds
using one training seed. \textbf{Bold} and \uline{underline} indicate the best
and second-best results, excluding the full fine-tuning references (gray).
\item[*] Our reimplementation under this protocol.
\end{tablenotes}
\end{threeparttable}
\end{table*}
\section{Experimental Setup}
\subsection{Datasets and preparation}
We evaluated two micro-Doppler HAR datasets with complementary acquisition settings 
and overlapping activity sets. Their differences in radar hardware, participant populations, 
and sensing geometry enabled evaluation under 
several deployment shifts.

\textbf{Glasgow}~\cite{glasgow,fioranelli2019healthcare} contained six activities:
walking, sitting down, standing up, picking up an object, drinking, and falling.
Recordings used an Ancortek frequency-modulated continuous-wave (FMCW) radar at
5.8\,GHz, with 400\,MHz bandwidth and a 1\,ms chirp. Seven acquisition campaigns
covered laboratory and clinical sites, with participants aged 21--98.

\textbf{MAD}~\cite{mad} contained ten activities, counting arbitrary-trajectory
walking separately. Recordings used an Ancortek SDR-980AD2 radar at 9.8\,GHz
with 400\,MHz bandwidth. The dataset included 100 laboratory subjects aged
18--60, balanced across the 18--40/41--60 age groups and gender. Activities were
recorded at labeled aspect angles.

For Glasgow, we split each 10\,s walking clip into two 5\,s windows to standardize
duration. Glasgow reuses participant identifiers across acquisition campaigns.
We conservatively grouped folds by identifier, keeping distinct participants
together when identifiers coincided while preserving subject disjointness.
For MAD, the two receive channels were retained as separate samples and grouped
by subject in all splits. This yielded 2,081 Glasgow and 17,716 MAD spectrograms.

Each spectrogram was log-compressed, resized to $224 \times 224$, replicated to
three channels, min--max normalized per sample, and standardized with ImageNet
statistics. We used no data augmentation and disabled random cropping and
flipping in the reference SelaFD implementation. All methods therefore received
identical preprocessed inputs. Because walking was over-represented after
segmentation, we also computed macro-F1 for every experiment. It preserved every
ranking in every table (mean gap 1.3\,pp). We report accuracy and release the
full macro-F1 tables with the code.

\subsection{Evaluation protocol}\label{sec:protocol}
\textbf{In-distribution.} We used five-fold subject-disjoint cross-validation,
with stratified grouping by participant identifier. We report mean accuracy
and standard deviation across folds.

\textbf{Out-of-distribution.} Let $\mathcal D_s$ and $\mathcal D_t$
denote the source and target datasets. Adapted encoders were fine-tuned
on $\mathcal D_s$. All encoders were then frozen for target evaluation.
Five-fold stratified group cross-validation partitioned $\mathcal D_t$
into target-training and target-test sets,
$\mathcal D_{t,k}^{\mathrm{train}}$ and
$\mathcal D_{t,k}^{\mathrm{test}}$, with disjoint subjects in each fold $k$.
For each support draw $j$, we sampled
$\mathcal S_{k,j}\subset\mathcal D_{t,k}^{\mathrm{train}}$
containing ten labeled examples per class. A linear probe was fit
on the support embeddings and evaluated on
$\mathcal D_{t,k}^{\mathrm{test}}$.\footnote{%
Direct source-head transfer failed on the harder shifts.
The best zero-shot accuracy was 47.2\% on cross-orientation,
versus 83.6\% at 10-shot. No method exceeded the 29.4\% majority
rate on cross-dataset. Cross-age zero-shot preserved the ranking.
Full results are released.}
We used ten support draws per fold, identical across methods.

For each source fine-tuning seed, accuracy was averaged over five folds
and ten support draws per fold. Table~\ref{tab:ood} reports the mean
and standard deviation across three such seed means. Frozen baselines
were averaged over the same folds and support draws.

\textbf{Cross-age (demographic shift).} We trained on Glasgow D1--D5
(60 subjects, ages 21--44) and evaluated on the older D6--D7 cohort
(35 subjects, ages 45--98). Five younger participants were excluded to make the
cohorts age-disjoint. Evaluation included five classes, excluding falls, which
were not recorded from older participants. This was the mildest shift, with
radar, geometry, and acquisition protocol fixed.

\textbf{Cross-orientation (sensing-geometry shift).} We trained on MAD's
``toward'' and ``away'' subsets and evaluated on the held-out ``right-to-left''
subset, keeping the subject pool fixed. Evaluation included nine classes,
excluding arbitrary-trajectory walking, which was available only in one aspect
subcategory. Tangential motion compresses the radial Doppler spread, producing
target signatures unlike those seen during source training.

\textbf{Cross-dataset (compound shift).} We trained on Glasgow and evaluated on
MAD over the five shared activities, excluding drinking. This combined changes
in radar, carrier frequency, spectrogram computation, environment, protocol,
and cohort.

\subsection{Baselines}
\textbf{Full fine-tuning.} We trained all parameters of ResNet50
(torchvision \texttt{IMAGENET1K\_V2} weights~\cite{resnet}) and DINOv3-S.
This compared a conventional CNN baseline with the backbone used by our method.

\textbf{Frozen features.} We evaluated supervised ViT-S/B
(AugReg, ImageNet-21k$\rightarrow$1k~\cite{augreg}) and DINOv3-S/B.
Standardized CLS embeddings were classified using $\ell_2$-regularized
multinomial logistic regression (scikit-learn, lbfgs, $C=1$). The same probe
was used for few-shot target readout, with feature standardization fit only
on the support set.

\textbf{Prior PEFT state of the art.} We reimplemented SelaFD~\cite{selafd}
under our protocol (marked $^{*}$) and constructed a Small variant for
size-matched comparison.

\textbf{Attention-only LoRA.} We adapted the supervised ViT using the standard
attention-only recipe, which also formed the reference configuration for our
ablation.

Out-of-distribution comparisons used Small models throughout to target
deployment on constrained hardware.

\subsection{Implementation details}
All fine-tuned methods were trained for 60 epochs with AdamW and cross-entropy
loss. We used a learning rate of $10^{-4}$, weight decay of $10^{-4}$, batch
size 16, and a cosine learning-rate schedule. All experiments were conducted on a laptop equipped with an NVIDIA
RTX PRO 1000 Blackwell Generation Laptop GPU.

\section{Results and Discussion}
\subsection{In-distribution (Table~\ref{tab:indist})}
Full fine-tuning achieved the highest in-distribution accuracy (94.6/\allowbreak{}97.9\% on
Glasgow/MAD), with similar results for ResNet50 and DINOv3-S. Among PEFT and
frozen methods, DINOv3-PiSSA-B and -S ranked first and second on both datasets.
The Small variant achieved slightly higher mean accuracy than Base-size SelaFD 
(+0.73/+0.61 percentage points), with over 20× fewer trainable parameters and 
4.5× fewer total parameters. Against size-matched SelaFD-S, its margins were 2.0/1.5\,pp.

Frozen probing was substantially weaker, particularly on ten-class MAD
(63--75\%), showing the importance of adapting pretrained representations to
micro-Doppler recognition. DINOv3-PiSSA approached full fine-tuning accuracy
while training only 3.0\% and 1.5\% of total model parameters for the Small and
Base variants, respectively. SelaFD's originally reported
96.6\% on Glasgow~\cite{selafd} came from a single random split and seed. Under
our subject-disjoint cross-validation, with augmentation disabled for input
parity, its reimplementation attained 92.1\%.
\begin{table}[t]
\caption{10-Shot Transfer Accuracy under Distribution Shift (\%)}
\label{tab:ood}
\centering
\setlength{\tabcolsep}{3pt}
\begin{threeparttable}
\begin{tabular*}{\columnwidth}{@{\extracolsep{\fill}}lccc@{}}
\toprule
Method & \shortstack{Cross-\\age} & \shortstack{Cross-\\orient.} & \shortstack{Cross-\\dataset} \\
\midrule
\multicolumn{4}{l}{\textit{Full fine-tuning}}\\
ResNet50 & \snd{79.12 (0.32)} & 79.37 (1.66) & 33.07 (3.71)\\
DINOv3-S (full FT) & \best{82.10 (0.73)} & \snd{81.35 (0.69)} & 30.93 (2.97)\\
\addlinespace
\multicolumn{4}{l}{\textit{Linear probing on frozen models}}\\
ViT-S (IN) & 66.25 & 29.29 & 55.28\\
DINOv3-S   & 64.68 & 38.22 & \snd{63.93}\\
\addlinespace
\multicolumn{4}{l}{\textit{PEFT-based}}\\
SelaFD-S$^{*}$ & 75.34 (1.01) & 77.00 (0.21) & 57.98 (0.79)\\
ViT-S + LoRA & 77.22 (0.78) & 75.83 (0.83) & 57.31 (0.92)\\
DINOv3-PiSSA-S & 77.46 (0.29) & \best{83.56 (1.02)} & \best{66.92 (0.38)}\\
\bottomrule
\end{tabular*}
\begin{tablenotes}[flushleft]\normalsize
\item Ten labeled target examples per class, with subject-disjoint support and
test sets. Values are mean accuracy (standard deviation) across three source
fine-tuning seeds. Frozen rows report mean accuracy only. See
Sec.~\ref{sec:protocol} for averaging over folds and support draws.
\textbf{Bold} and \uline{underline} indicate the best and second-best results
across all methods.
\item[*] Our reimplementation under this protocol.
\end{tablenotes}
\end{threeparttable}
\end{table}
\subsection{Out-of-distribution (Table~\ref{tab:ood})}\label{sec:ood}
The in-distribution ranking did not persist under shift. PEFT was the only tested adaptation family to avoid a sharp
performance collapse across all three shifts. Full fine-tuning suffered
negative transfer on cross-dataset: both fully fine-tuned architectures reached
only 30.9--33.1\%, below the frozen baselines (55--64\%). For the same DINOv3-S backbone,
 source fine-tuning reduced accuracy from 63.9\% to 30.9\%. Frozen
features showed the opposite weakness on cross-orien\-tation (29--38\%), where
source-adapted representations performed substantially better. With the same
DINOv3-S backbone, the three adaptation strategies spanned 45\,pp on
cross-orien\-tation and 36\,pp on cross-dataset.

DINOv3-PiSSA-S achieved the highest accuracy on the two harder shifts
(83.6/\allowbreak{}66.9\%). Full fine-tuning remained strongest on cross-age (82.1\% versus
77.5\% with the same backbone), but its advantage reversed across the three
shifts ($+4.6$, $-2.2$, $-36.0$\,pp). Our margins over SelaFD-S increased
correspondingly ($+2.1$, $+6.6$, $+8.9$\,pp). On cross-dataset, our method also
surpassed the frozen backbone, whereas the ImageNet-based adapters remained
below it. Together with Table~\ref{tab:indist}, these results placed our method
first among PEFT methods in every evaluated setting.

The contrasting failure patterns explain why in-distribution accuracy alone
was insufficient for selecting an adaptation strategy. Full fine-tuning
performed well under the milder age shift, but its source-domain advantage did
not persist under compound shift. Frozen representations retained useful
cross-dataset features but adapted poorly to the orientation change. PEFT's
performance across both settings is consistent with a balance between learning
source-task features and preserving transferable pretrained representations.
\begin{table}[t]
\caption{Pretraining and Adapter-Recipe Ablation under Distribution Shift (\%)}
\label{tab:abl}
\centering
\setlength{\tabcolsep}{2pt}
\begin{threeparttable}
\begin{tabular}{@{}lcccc@{}}
\toprule
 & \multicolumn{2}{c}{Cross-orientation} & \multicolumn{2}{c}{Cross-dataset}\\
\cmidrule(lr){2-3}\cmidrule(lr){4-5}
 & ViT-S (IN) & DINOv3-S & ViT-S (IN) & DINOv3-S \\
\midrule
LoRA (attn) & 75.83 (0.83) & 80.49 (1.11) & 57.31 (0.92) & \snd{65.54 (1.73)}\\
PiSSA (all) & \snd{81.19 (1.50)} & \best{83.56 (1.02)} & 58.15 (1.27) & \best{66.92 (0.38)}\\
\bottomrule
\end{tabular}
\begin{tablenotes}[flushleft]\normalsize
\item Attention-only LoRA and all-layer PiSSA applied to supervised ViT-S and
DINOv3-S. Evaluation followed Table~\ref{tab:ood}'s 10-shot protocol.
Values are mean accuracy (standard deviation) across three source fine-tuning
seeds. \textbf{Bold} and \uline{underline} indicate the best and second-best of
the four configurations within each shift.
\end{tablenotes}
\end{threeparttable}
\end{table}
\subsection{Ablation (Table~\ref{tab:abl})}\label{sec:ablation}
The ablation separated pretraining (supervised ViT-S versus DINOv3-S) from
adapter recipe (attention-only LoRA versus all-layer PiSSA). Both changes
improved mean accuracy on both harder shifts. Averaged over the adapter recipes, 
switching to DINOv3 pretraining improved accuracy by $+8.5$\,pp on cross-dataset, 
compared with $+1.1$\,pp
for the recipe. On cross-orien\-tation, their contributions were more comparable
($+3.5$ and $+4.2$\,pp, respectively). The combined changes improved accuracy by 9.6 pp 
on cross-dataset and 7.7 pp on cross-orientation 
over supervised ViT-S with attention-only LoRA.

The backbone's dominant contribution on cross-dataset highlighted the
importance of pretrained representation choice under compound shift,
consistent with the hypothesis in Sec.~\ref{sec:backbone}. On cross-orien\-tation,
the larger contribution of the adapter recipe showed that effective adaptation
also mattered: the strongest pretrained backbone benefited further from
all-layer PiSSA.

The recipe comparison combined initialization and layer coverage. A separate
controlled $2\times2$ ablation, released with the code, found that attention-only
PiSSA and all-layer LoRA both underperformed attention-only LoRA. Their
combination was necessary for the observed improvement, supporting the joint
use of principal-direction initialization and all-layer adaptation.
\section{Conclusion}
\textbf{RQ1} asked which adaptation regime best preserved in-distribution
accuracy and few-shot transfer. Full fine-tuning led
in-distribution (94.6/\allowbreak{}97.9\%) but DINOv3-S fell to 30.9\% on cross-dataset.
Frozen DINOv3-S retained 63.9\% there but fell to 38.2\% on cross-orien\-tation.
PEFT avoided both failures, reaching 83.6/\allowbreak{}66.9\% on
cross-orien\-tation/cross-dataset. \textbf{RQ2} asked whether self-supervised image
backbones could improve on supervised ImageNet PEFT. DINOv3-PiSSA surpassed
SelaFD-S by 2.1/6.6/8.9\,pp across
the three shifts and Base-size SelaFD by 0.73/0.61\,pp in-distribution. The
Small variant trained 3\% of its backbone and used $4.5\times$ fewer total
parameters than SelaFD. Under compound shift, the backbone contributed
$+8.5$\,pp versus $+1.1$\,pp for the adapter recipe. \\
\textbf{Limitations and future work.} Cross-dataset transfer remained challenging
despite substantial gains: DINOv3-PiSSA reached 66.9\% accuracy, compared with
30.9\% for full fine-tuning of the same backbone. Strong performance on the two
harder shifts still required a small labeled target support set (10 examples
per class). All tested backbones were pretrained on natural images.
Audio-spectrogram pretraining or a brief micro-Doppler self-supervised phase
could narrow this gap. \\
We release code, protocol, and preprocessing at
\url{https://github.com/KeryanChelouche/udoppler-ssl}. Glasgow is publicly
available~\cite{glasgow}; Access to MAD~\cite{mad}
must be requested at
\href{mailto:olivier.romain@cyu.fr}{olivier.romain@cyu.fr}.

\clearpage
\section{Acknowledgment}
This work was supported by the French National Research Agency 
(ANR) under grant 24-LCV2-0020-01 /SmartGaitLab. \\
\textbf{AI use disclosure} The authors used ChatGPT (OpenAI) for manuscript review and editing, 
and Claude Code (Anthropic) for assistance with developing the experimental code. 
The authors reviewed and revised the resulting text and code and take full responsibility for the final manuscript, 
implementation, and reported results.
\section{Compliance with Ethical Standards}
This study used existing recordings from Glasgow~\cite{glasgow} and
MAD~\cite{mad}, without new participant recruitment or data collection.
Glasgow was released under CC BY 4.0, with consent documentation permitting
future research using anonymized data. MAD acquisition was approved by the
Research Ethics Committee of CY Cergy Paris Université (CER-CY, protocol
202307-004, 11 October 2023). Participants provided written informed consent before recording,
including permission for future research use of their data~\cite{mad}.

\bibliographystyle{IEEEbib}
\bibliography{refs}

@misc{who_falls,
  author = {{World Health Organization}},
  title = {{Falls}},
  howpublished = {Fact sheet},
  month = {Apr.~26,},
  year = {2021},
  note = {[Online]. Available: \url{https://www.who.int/news-room/fact-sheets/detail/falls}},
}

@article{gurbuz2019,
  author = {S. Z. Gurbuz and M. G. Amin},
  title = {{Radar-based human-motion recognition with deep learning: Promising applications for indoor monitoring}},
  journal = {IEEE Signal Process. Mag.},
  volume = {36},
  number = {4},
  pages = {16--28},
  year = {2019},
  note = {DOI: \href{https://doi.org/10.1109/MSP.2018.2890128}{\nolinkurl{10.1109/MSP.2018.2890128}}},
}

@article{radarreview,
  author = {M. T. Buyukakkaslar and M. A. Erturk and M. A. Aydin},
  title = {{A review on radar-based human detection techniques}},
  journal = {Sensors},
  volume = {24},
  number = {17},
  year = {2024},
  note = {Art. no. 5709, DOI: \href{https://doi.org/10.3390/s24175709}{\nolinkurl{10.3390/s24175709}}},
}

@article{kim2016,
  author = {Y. Kim and T. Moon},
  title = {{Human detection and activity classification based on micro-Doppler signatures using deep convolutional neural networks}},
  journal = {IEEE Geosci. Remote Sens. Lett.},
  volume = {13},
  number = {1},
  pages = {8--12},
  year = {2016},
  note = {DOI: \href{https://doi.org/10.1109/LGRS.2015.2491329}{\nolinkurl{10.1109/LGRS.2015.2491329}}},
}

@article{li2020,
  author = {H. Li and A. Shrestha and H. Heidari and {Le Kernec}, J. and F. Fioranelli},
  title = {{Bi-LSTM network for multimodal continuous human activity recognition and fall detection}},
  journal = {IEEE Sensors J.},
  volume = {20},
  number = {3},
  pages = {1191--1201},
  year = {2020},
  note = {DOI: \href{https://doi.org/10.1109/JSEN.2019.2946095}{\nolinkurl{10.1109/JSEN.2019.2946095}}},
}

@article{diatradharnet,
  author = {M. Chakraborty and H. C. Kumawat and S. V. Dhavale and {A. B. Raj A.}},
  title = {{DIAT-RadHARNet: A lightweight DCNN for radar based classification of human suspicious activities}},
  journal = {IEEE Trans. Instrum. Meas.},
  volume = {71},
  year = {2022},
  note = {Art. no. 2505210, DOI: \href{https://doi.org/10.1109/TIM.2022.3154832}{\nolinkurl{10.1109/TIM.2022.3154832}}},
}

@inproceedings{selafd,
  author = {Y. Wang and Y. Wang and C. Xu and S. Yao and Q. Wu},
  title = {{SelaFD: Seamless adaptation of vision transformer fine-tuning for radar-based human activity recognition}},
  booktitle = {Proc. IEEE Int. Conf. Acoust., Speech, Signal Process. (ICASSP)},
  year = {2025},
  pages = {1--5},
  note = {DOI: \href{https://doi.org/10.1109/ICASSP49660.2025.10888271}{\nolinkurl{10.1109/ICASSP49660.2025.10888271}}},
}

@misc{picard2021,
  author = {D. Picard},
  title = {{Torch.manual\_seed(3407) is all you need: On the influence of random seeds in deep learning architectures for computer vision}},
  year = {2021},
  howpublished = {arXiv:2109.08203},
}

@inproceedings{bouthillier2021,
  author = {X. Bouthillier and others},
  title = {{Accounting for variance in machine learning benchmarks}},
  booktitle = {Proc. MLSys},
  year = {2021},
}

@inproceedings{recht2019,
  author = {B. Recht and R. Roelofs and L. Schmidt and V. Shankar},
  title = {{Do ImageNet classifiers generalize to ImageNet?}},
  booktitle = {Proc. ICML},
  year = {2019},
}

@inproceedings{torralba2011,
  author = {A. Torralba and A. A. Efros},
  title = {{Unbiased look at dataset bias}},
  booktitle = {Proc. CVPR},
  year = {2011},
  pages = {1521--1528},
}

@article{saeb2017,
  author = {S. Saeb and L. Lonini and A. Jayaraman and D. C. Mohr and K. P. Kording},
  title = {{The need to approximate the use-case in clinical machine learning}},
  journal = {GigaScience},
  volume = {6},
  number = {5},
  year = {2017},
  note = {DOI: \href{https://doi.org/10.1093/gigascience/gix019}{\nolinkurl{10.1093/gigascience/gix019}}},
}

@article{shah2022,
  author = {S. A. Shah and A. Tahir and {Le Kernec}, J. and A. Zoha and F. Fioranelli},
  title = {{Data portability for activities of daily living and fall detection in different environments using radar micro-Doppler}},
  journal = {Neural Comput. Appl.},
  volume = {34},
  number = {10},
  pages = {7933--7953},
  year = {2022},
  note = {DOI: \href{https://doi.org/10.1007/s00521-022-06886-2}{\nolinkurl{10.1007/s00521-022-06886-2}}},
}

@article{chen2003,
  author = {V. C. Chen and F. Li and S.-S. Ho and H. Wechsler},
  title = {{Analysis of micro-Doppler signatures}},
  journal = {IEE Proc. Radar Sonar Navig.},
  volume = {150},
  number = {4},
  pages = {271--276},
  year = {2003},
  note = {DOI: \href{https://doi.org/10.1049/ip-rsn:20030743}{\nolinkurl{10.1049/ip-rsn:20030743}}},
}

@inproceedings{caron2021,
  author = {M. Caron and H. Touvron and I. Misra and H. J\'egou and J. Mairal and P. Bojanowski and A. Joulin},
  title = {{Emerging properties in self-supervised vision transformers}},
  booktitle = {Proc. ICCV},
  year = {2021},
}

@article{oquab2024,
  author = {M. Oquab and others},
  title = {{DINOv2: Learning robust visual features without supervision}},
  journal = {Trans. Mach. Learn. Res.},
  year = {2024},
  note = {arXiv:2304.07193},
}

@unpublished{wifidino,
  author = {{Anonymous}},
  title = {{Frozen vision foundation models narrow the sensor-shift gap in non-visual spectrogram recognition}},
  note = {Submitted to \textit{Trans. Mach. Learn. Res.}, unpublished. [Online]. Available: \url{https://openreview.net/forum?id=akIq8wHuUD}},
}

@misc{dinov3,
  author = {O. Sim\'eoni and others},
  title = {{DINOv3}},
  year = {2025},
  howpublished = {arXiv:2508.10104},
  note = {[Online]. Available: \url{https://arxiv.org/abs/2508.10104}},
}

@inproceedings{lora,
  author = {E. J. Hu and others},
  title = {{LoRA: Low-rank adaptation of large language models}},
  booktitle = {Proc. ICLR},
  year = {2022},
}

@inproceedings{pissa,
  author = {F. Meng and Z. Wang and M. Zhang},
  title = {{PiSSA: Principal singular values and singular vectors adaptation of large language models}},
  booktitle = {Proc. NeurIPS},
  year = {2024},
}

@misc{glasgow,
  author = {F. Fioranelli and S. A. Shah and H. Li and A. Shrestha and S. Yang and {Le Kernec}, J.},
  title = {{Radar signatures of human activities}},
  howpublished = {University of Glasgow, dataset},
  year = {2019},
  note = {DOI: \href{https://doi.org/10.5525/gla.researchdata.848}{\nolinkurl{10.5525/gla.researchdata.848}}},
}

@phdthesis{mad,
  author = {C. B\'eranger},
  title = {{Gait recognition and fall prediction with deep-learning on micro-Doppler signatures}},
  school = {ETIS UMR 8051, CY Cergy Paris Univ.},
  address = {Cergy, France},
  year = {2026},
  note = {[Online]. Available: \url{https://theses.fr/2026CYUN0004}},
}

@inproceedings{resnet,
  author = {K. He and X. Zhang and S. Ren and J. Sun},
  title = {{Deep residual learning for image recognition}},
  booktitle = {Proc. CVPR},
  year = {2016},
  pages = {770--778},
}

@article{augreg,
  author = {A. Steiner and A. Kolesnikov and X. Zhai and R. Wightman and J. Uszkoreit and L. Beyer},
  title = {{How to train your ViT? Data, augmentation, and regularization in vision transformers}},
  journal = {Trans. Mach. Learn. Res.},
  year = {2022},
}

@article{fioranelli2019healthcare,
  author = {F. Fioranelli and S. A. Shah and H. Li and A. Shrestha and S. Yang and {Le Kernec}, J.},
  title = {{Radar sensing for healthcare}},
  journal = {Electronics Letters},
  volume = {55},
  number = {19},
  pages = {1022--1024},
  year = {2019},
  note = {DOI: \href{https://doi.org/10.1049/el.2019.2378}{\nolinkurl{10.1049/el.2019.2378}}},
}
\end{document}